\documentclass[letterpaper, 10 pt, journal, twoside]{IEEEtran} 
\usepackage{times}
\usepackage{amsfonts} 
\usepackage{graphicx}
\usepackage{comment}
\usepackage{wrapfig}
\usepackage{amsmath}
\usepackage{amssymb}
\usepackage{array}
\usepackage{wrapfig,lipsum,booktabs}
\usepackage{booktabs}
\usepackage{multirow}
\usepackage{subfig}
\usepackage{setspace}
\usepackage{xcolor}
\usepackage{array,longtable}

\usepackage{algorithm}
\usepackage{algorithmic}

\IEEEoverridecommandlockouts                              

\newcommand{\revision}[1]{{\color{black} #1}}

\usepackage[numbers]{natbib}
\usepackage{multicol}
\usepackage[bookmarks=true]{hyperref}

\begin{document}


\title{
On the Effectiveness of Retrieval, Alignment, and Replay in Manipulation
}

\author{Norman Di Palo$^{1}$ and Edward Johns$^{1}$%
\thanks{Manuscript received: August 21, 2023; Revised November 9, 2023; Accepted December 8, 2023 .}
\thanks{This paper was recommended for publication by Editor Aleksandra Faust upon evaluation of the Associate Editor and Reviewers' comments.
This work was supported by the Royal Academy of Engineering under the Research Fellowship Scheme.} 
\thanks{$^{1}$Norman Di Palo and Edward Johns are with the Robot Learning Lab at Imperial College London {\tt\footnotesize n.di-palo20@imperial.ac.uk}}%
\thanks{Digital Object Identifier (DOI): see top of this page.}
}



%

\maketitle

\begin{abstract}

   Imitation learning with visual observations is notoriously inefficient when addressed with end-to-end behavioural cloning methods. In this paper, we explore an alternative paradigm which decomposes reasoning into three phases. First, a \textit{retrieval} phase, which informs the robot \textit{what} it can do with an object. Second, an \textit{alignment} phase, which informs the robot \textit{where} to interact with the object. And third, a \textit{replay} phase, which informs the robot \textit{how} to interact with the object. Through a series of real-world experiments on everyday tasks, such as grasping, pouring, and inserting objects, we show that this decomposition brings unprecedented learning efficiency, and effective inter- and intra-class generalisation. Videos are available at \href{https://www.robot-learning.uk/retrieval-alignment-replay}{https://www.robot-learning.uk/retrieval-alignment-replay}.
\end{abstract}

\begin{IEEEkeywords}
Deep Learning in Grasping and Manipulation, Imitation Learning, Learning from Demonstration\end{IEEEkeywords}

\IEEEpeerreviewmaketitle

\begin{figure}[t!]
    \centering
    \includegraphics[width=0.43\textwidth]{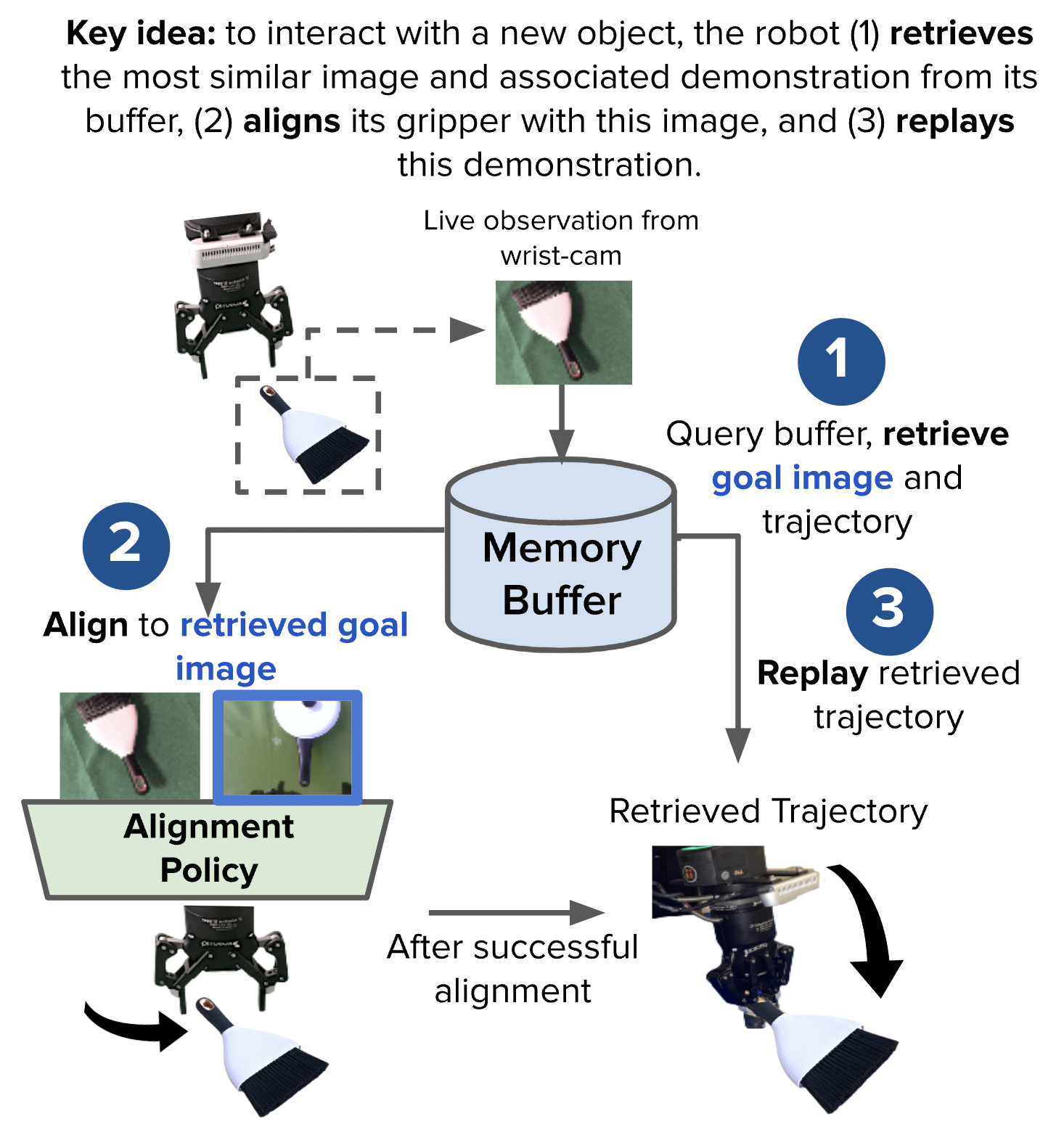}
    \caption{An overview of the framework we study, showing the \textit{retrieval}, \textit{alignment}, and \textit{replay} phases. Together, these enable one-shot imitation learning without prior object knowledge, as well as generalisation to novel objects and novel classes.}
    \label{fig:brief}
\end{figure}

\section{Introduction}

\IEEEPARstart{I}{n} this paper, we study the problem of teaching a robot how to interact with a set of training objects, and then generalising these learned behaviours to novel objects and novel classes of objects. Today's dominant paradigm in recent literature is to address this with end-to-end behavioural cloning \cite{open_x_embodiment_rt_x_2023}. However, to generalise everyday manipulation skills to many different objects, such techniques require a very large number of human demonstrations, which is slow and expensive.

But as an alternative to monolithic, end-to-end control, we can decompose reasoning into three distinct, specialist modes of reasoning. Firstly, \textit{what} can a robot do with an object? Secondly, \textit{where} should a robot interact with an object? And thirdly, \textit{how} should a robot interact with an object? Our hypothesis is that this decomposition might be more optimal than expecting a single control policy to be able to reason simultaneously about all three modes.

To achieve these three modes of reasoning, we propose a new framework based on three principles: \textbf{retrieval, alignment, and replay}. When provided with an object during testing, these are then used as follows. First, to determine \textit{what} can be done with the object, the most visually similar training object to the observed object is \textit{retrieved} from a memory buffer of the training objects. Second, to determine \textit{where} the robot should interact with the object, a goal image representing the retrieved training object is used in a goal-conditioned \textit{alignment} policy, to align the end-effector appropriately with the object. Third, to determine \textit{how} the robot should interact with object, the robot \textit{replays} the demonstration velocities for that training object.

As such, our hypothesis is that, in the absence of an explicit demonstration for a novel test object, interacting in the same way as with the most visually similar training object, might be sufficient to interact with the test object. And through a series of real-world experiments, we demonstrate this to be true for a range of everyday tasks such as grasping, pouring, and inserting. Furthermore, we study whether our proposed decomposition into individual modes of reasoning, is optimal compared to methods which do decomposition but not retrieval (e.g. \cite{lee2020guided}), to methods which do retrieval but not decomposition (e.g. \cite{pari2021surprising}), and to methods which do neither decomposition nor retrieval (e.g. behavioural cloning \cite{open_x_embodiment_rt_x_2023}). Our conclusion, and the key takeaway message, is clear: decomposing into retrieval, alignment, and replay leads to an order of magnitude better learning efficiency compared to all these other alternative paradigms. Furthermore, \textbf{our framework only requires a wrist-mounted camera, without the need for either extrinsic external or intrinsic calibration, and begins learning \textit{tabula rasa}, with no previous knowledge of the objects required, such as CAD models}: \revision{our only assumption is that the relative pose between wrist-camera and end-effector remains fixed between train and test time, i.e. the camera is rigidly mounted to the robot}. 

\section{Related Work}
\label{sec:related}
Many recent methods that learn object manipulation from demonstrations, learn end-to-end control policies using behavioural cloning, which can be conditioned on a task identity or language command to enable generalisation to novel objects. However, such policies tend to require extensive datasets of human demonstrations \cite{open_x_embodiment_rt_x_2023}. In our work, we show that an explicit decomposition into \textit{retrieval}, \textit{alignment}, and \textit{replay}, results in a substantial improvement in data and time efficiency over end-to-end control methods.

Our work builds on top of Coarse-to-Fine Imitation Learning (C2F-IL) \cite{johns2021coarse}, which also proposes alignment and replay. However, C2F-IL learns tasks in isolation: it learns a single control policy for a single object, which does not generalise to novel objects. As such, we study a framework based on retrieval, which can then scale up and learn from many different demonstrations across many different objects, generalising to both novel objects and novel object classes, and becoming more and more capable with each new demonstration.

Other works have proposed a decomposition of the robot trajectory into a coarse, approaching stage and a fine, last-inch interaction \cite{ lee2020guided, wen2022you, borja2022affordance, vosylius2023start, vitiello2023one}. However, most have stronger constraints in their design compared to our proposed framework: \cite{wen2022you} needs synthetic visual data, and therefore models of the objects, to train a perception system in simulation. Our method starts learning \textit{tabula rasa}, and needs no prior knowledge of the tasks or objects, therefore making it substantially more general and deployable in everyday environments. \cite{borja2022affordance} also performs trajectory decomposition, but needs external, calibrated RGB-D cameras to identify affordances. \cite{vosylius2023few} frames robot manipulation entirely as a series of objects alignments, it however relies on point clouds and calibrated external cameras. Our method only needs a single uncalibrated wrist-camera, which is significantly easier to deploy in unstructured environments.

The use of retrieval for few-shot learning, especially in robotics, reinforcement learning and control, has also been investigated elsewhere \cite{du2023behavior, pari2021surprising}. \cite{pari2021surprising} directly retrieves the actions to execute, whilst \cite{du2023behavior} retrieves task-related data from an unlabelled buffer to train an end-to-end policy. We combine retrieval and decomposition in a novel way, through an interplay of retrieving goal observations and trajectories to execute and learning of an alignment policy through self-supervised learning. 

\section{Method}
\label{sec:method}

We aim to provide robots with the capability to both efficiently learn to interact with an object, and transfer the learned behaviour to novel objects. To achieve this, we propose a novel imitation learning framework based on an interplay between \textbf{retrieval} from a memory buffer of visual observations, a learned goal-conditioned visual \textbf{alignment} policy, and \textbf{replay} of demonstrated trajectories. Our method only needs an RGB-D wrist-camera, rigidly mounted to the robot's end-effector, without the need for any calibration. No prior knowledge of objects or tasks is needed, and no external camera is needed.


\subsection{Preliminaries}
\label{sec:preliminaries}
Unless explicitly stated otherwise, we use these terms with the following meanings: a \textit{trajectory} executed by the robot is composed of 3D linear and 3D angular velocities, expressed in the end-effector frame. A \textit{pose} refers to a 4D vector $x,y,z,\theta_z$ expressed in the world frame: it can describe either the pose of an object on a table, or the pose of the end-effector in the environment. A \textit{policy} is a function mapping visual inputs to action outputs. An \textit{action} is an end-effector velocity executed at each time-step. A \textit{task}, in this work, can be intuitively described as a verb, like \textit{grasp}. The same task can be performed on many objects (e.g. grasping a mug, grasping a teapot). The environment is a tabletop setting with an object the robot should interact with. All the tasks we consider involve executing full 6-DoF velocities during object interaction.

\begin{figure*}[t!]
    \centering
    \includegraphics[width=1.\textwidth]{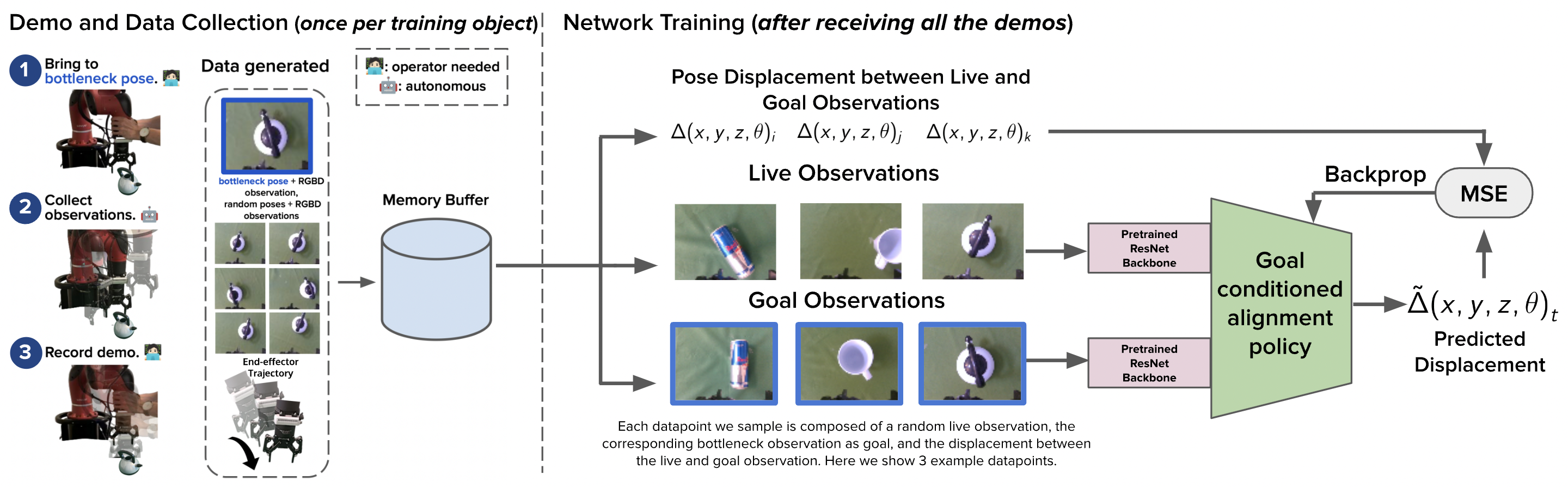}
    \caption{We collect a single demonstration per object following the steps in Sec. \ref{sec:c2f}. Through autonomously collected data, our method trains a goal-conditioned alignment policy, that will align the robot to the bottleneck pose before executing the trajectory. All collected data is stored in the memory buffer.}
    \label{fig:demo+train}
\end{figure*}

\subsection{An Overview of the Retrieval, Alignment and Replay Phases}
In this section, we provide a high-level overview of our framework and its workflow, as illustrated in Figure \ref{fig:brief}. 

When deployed during testing, the robot starts by obtaining a visual observation of the environment with its wrist-camera. To understand \textit{what} to do with the object it has observed, it queries an external memory buffer with the visual observation. The memory buffer \textbf{retrieves} \textit{where} to interact with the object in the form of a goal visual observation, and \textit{how} to interact with the object as a trajectory, expressed in the end-effector's frame. This retrieved data is used to guide the robot's behaviour as follows.

The goal of the robot, after retrieving this information from the buffer, is simple: to move its end-effector so that the live observation is \textbf{aligned} with the retrieved goal observation. This is done via visual servoing through a learned, goal-conditioned visual alignment policy. Once the alignment is completed, the robot \textbf{replays} the retrieved trajectory by executing the trajectory of end-effector velocities. This allows for successful interaction with new objects, under the hypothesis that objects which are visually similar should be interacted with in the same way, which we later validate experimentally.

The following subsections describe in more detail how each part is implemented.

\subsection{Understanding the Retrieval, Alignment, and Replay Phases}
\label{sec:c2f}

The order in which the phases are executed during deployment is \textit{retireval}, followed by \textit{alignment}, followed by  \textit{replay}. However, for clarity, we will now adopt the reverse order to describe them. This will allow us to first start with simple concepts, and to later describe the overall framework, providing the reader with a more intuitive sequence of ideas. The \textit{alignment} and \textit{replay} phases are inspired by Coarse-to-Fine Imitation Learning (C2F-IL) \cite{johns2021coarse}, which we now build upon using \textit{retrieval} to allow for generalisation to novel objects, and scalability by allowing for data from multiple demonstrations and across multiple tasks to be absorbed by the robot.

\subsubsection{\textbf{The Replay Phase: Recording and replaying demonstrated trajectories}}
\label{sec:how}

To teach the robot how to interact with an object, the human operator manoeuvres the end-effector to provide a demonstration, e.g. with kinesthetic teaching. The end-effector starts from a pose $b^O$, chosen by the operator and expressed in the imaginary object frame $O$, that we call the \textit{bottleneck pose}. The only requirement is that the object must be visible by the wrist-camera from the bottleneck pose. The demonstration is recorded as a series of 3D linear and 3D angular velocities of the end-effector expressed in the end-effector frame $E$, which we call a \textit{trajectory} $ s = [\mathcal{V}^E_1, \dots, \mathcal{V}^E_{T}]$, where $\mathcal{V} = [v_x, v_y, v_z, w_x, w_y, w_z]$.

We can imagine the bottleneck pose being rigidly attached to the imaginary object frame. With $W$ being the world frame, when we move the object, $b^W$ moves rigidly with it, while $b^O$ stays constant. If the end-effector is re-aligned to $b^W$ when the object is moved, re-executing the end-effector velocities of the recorded trajectory would suffice in solving the task again \cite{johns2021coarse}. Therefore, the learning process can focus on reaching $b^W$ again for novel poses of the object. The \textit{replay} phase is hence the execution of a recorded trajectory. We store all the trajectories demonstrated during training in our memory buffer as $[s_1, \dots, s_{D}]$, where $D$ is the total number of demonstrations.

\subsubsection{\textbf{The Alignment Phase: Learning to align to the bottleneck again}}
\label{sec:where}
Before the demonstration starts, the operator aligns the end-effector to $b^O$, which is wherever the operator plans to start recording the trajectory from. In this configuration, the robot can compute the exact pose of $b^W$ through forward kinematics. The goal of the \textit{alignment} phase is reaching $b^W$ again at test time, when its location cannot be simply computed from forward kinematics (because the object has now moved), and thus must be predicted from visual observations.

Our approach is to train a goal-conditioned visual alignment network that, guided by the wrist-camera observations, aligns the end-effector to $b^W$. The network receives, as input, the live observation of the wrist-camera, and an observation recorded at $b^W$: its goal is to align the former to the latter by moving the end-effector, and hence, the wrist-camera. To train this neural network, we need a dataset of visual observations of the object from many different poses, and the relative pose displacement that would align the end-effector to $b^W$ from each pose.

We gather this dataset as follows. We store the \textit{bottleneck visual observation} $o_b$ taken from $b^W$ and later use it as the goal observation when training the network. The robot then moves in the volume above $b^W$, reaching new random poses $p_i^W$ and collecting visual observations $o_i$ from its wrist-camera. Knowing both $p_i^W$ and $b^W$ through forward kinematics, it can compute the pose displacement $d_i^E$ in end-effector frame $E$ that would bring the end-effector back to $b^W$, aligning $o_i$ to $o_b$. All this data is added to the memory buffer. Once this autonomous process is complete, the operator can manually manoeuvre the end-effector to provide the demonstration as in Sec. \ref{sec:how}. As shown in \cite{johns2021coarse}, this enables a task to be learned from just a single demonstration. This process is illustrated in Figure \ref{fig:demo+train}, left side.

Once all the desired $D$ demonstrations have been collected, across a number of objects and tasks, our method can train the goal-conditioned visual alignment network through regression, using the data stored in the memory buffer. As illustrated in Figure \ref{fig:demo+train} right side, we sample pairs of $o_b, o_i$ collected from the same object, along with the pose displacement $d_i^E$. We use the former as inputs to the network, and the latter as the label, optimising the weights through backpropagation and gradient descent using an MSE loss.

\begin{figure*}[t!]
    \centering
    \includegraphics[width=.8\textwidth]{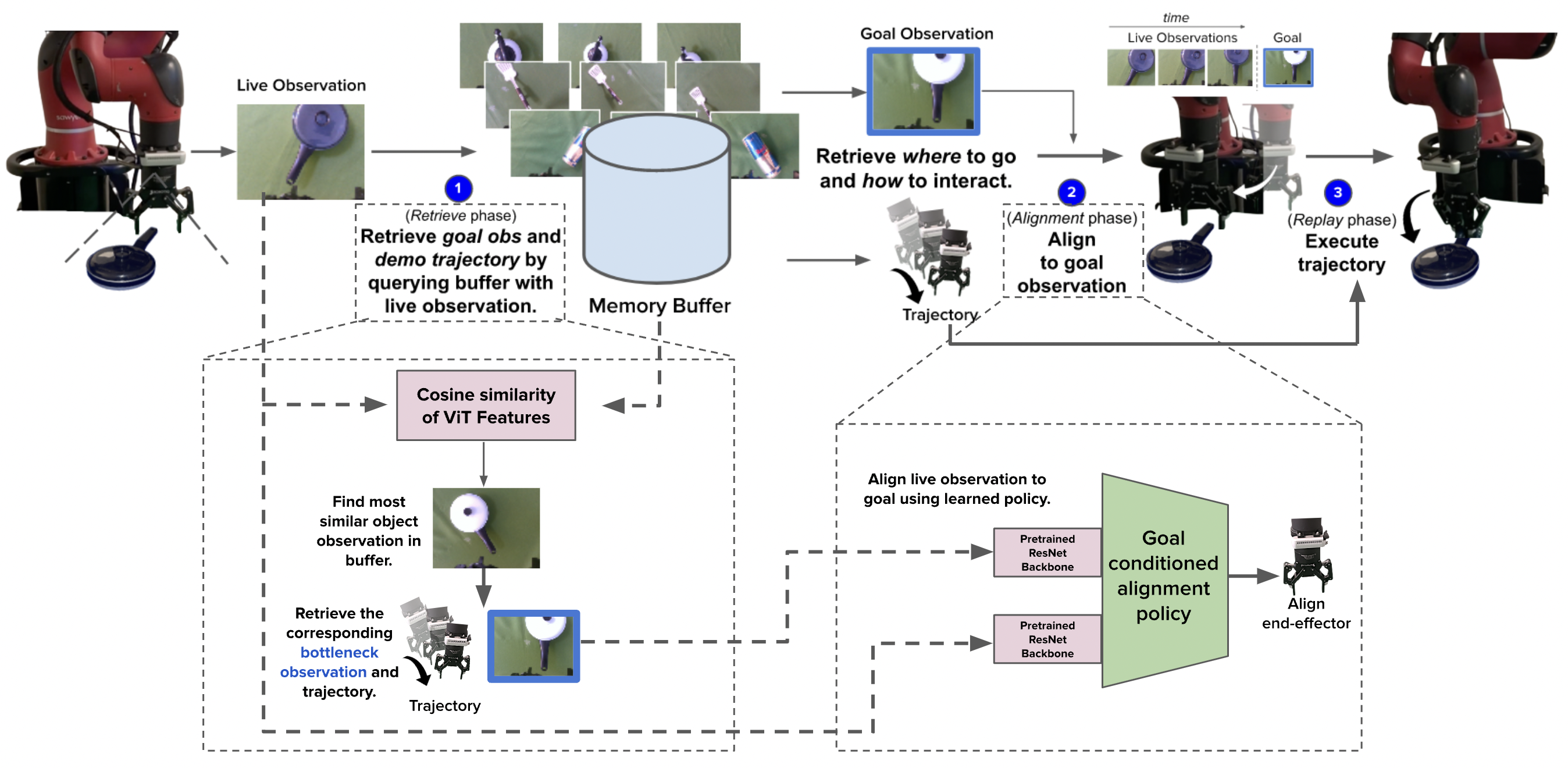}
    \caption{Overall illustration of our framework. Upon observing a new object, the agent visually compares it to other objects it observed during training. After finding the most similar object, the agent retrieves a goal observation of that object to guide the visual alignment phase (\textit{where} to interact), and also a trajectory to execute once the alignment is complete (\textit{how} to interact).}
    \label{fig:key-idea}
     \label{fig:retrieval}
\end{figure*}

At test time, when the object is in a different configuration and $b^W$ is unknown, we feed the visual input from the wrist-camera to the goal-conditioned visual alignment model along with the goal observation to align to, which will then predict the pose displacement $d_t^E$ between $p_t^W$, the current pose of the end-effector, and $b^W$: moving the end-effector by that pose will align it to $b^W$, so that the \textit{replay} phase can then be executed. This process constitutes the \textit{alignment} phase.

But, when at test time, and the robot observes a novel object, how does the robot obtain a goal observation to guide the goal-conditioned visual alignment network for the \textit{alignment} phase? Similarly, how does the robot then decide which trajectory to execute during the \textit{replay} phase? The robot must autonomously obtain this information guided only by the visual inputs it receives.  This constitutes the \textit{retrieval} phase, as now described.

\subsubsection{\textbf{The Retrieval Phase: Goal image and trajectory retrieval from the memory buffer}}
\label{sec:what}
After the training stage, where the robot receives a demonstration per each object in the dataset, our robot is equipped with an external memory buffer which stores the following: 1) the $D$ trajectories recorded by the operator, 2) the $D$ bottleneck visual observations $o_{b,d}$ used as goal observations, and 3) the observations $o_{i,d}$ collected for each object at random poses by the wrist-camera, with $0 \leq d < D, d \in \mathbb{N}$, $0 \leq i < I, i \in \mathbb{N}$, $I$ being the total number of observations collected per object, and $o_{i,d}$ therefore being the $i$-th observation of the $d$-th object.

After receiving the visual observation of an object the robot should interact with, the goal of the \textit{retrieval} phase is twofold: 1) to retrieve one of the $D$ \textit{bottleneck visual observations} to use as the goal for the goal-conditioned visual alignment network, guiding the \textit{alignment} phase, and 2) to retrieve one of the $D$ trajectories that, if executed during the \textit{replay} phase, will lead to a correct interaction with the object. The visual aspect of the novel object should therefore guide the choice of \textit{where} to align with the object, and \textit{how} to interact with it.

The robot obtains this information by querying the memory buffer with the live observation. Our framework uses a similarity function $g(o_{live}, o_{i,d}) \rightarrow [-1, 1]$ computed between the current live observation and all observations stored in the buffer. It firstly computes $o_{j,n} = \text{argmax}_{o_{i,d}} g(o_{live},o_{i,d})$, and then, it retrieves both the corresponding \textit{bottleneck visual observation} $o_{b,n}$, and the corresponding trajectory $s_{n}$. Intuitively, replicating the alignment and interaction of the most visually similar object in the buffer studies our hypothesis that \textit{visually similar objects should be interacted with in the same way}.

The recent computer vision literature proposes a variety of techniques to extract useful representations from visual inputs to map them to smaller dimensional embedding vectors, which capture semantic information of the image, where it is more efficient and effective to compute distances using well known mathematical functions. In our case, the similarity function is computed as the cosine similarity between the visual features of the RGB channels of the two observations extracted from a Vision Transformer (ViT) \cite{dosovitskiy2021an} trained using DINO \cite{caron2021emerging}. The literature has demonstrated how these features capture both geometric and semantic information on both a local and global scale, generalising between similar objects of the same or different classes \cite{amir2021deep}. We extract these features for all the observations in the buffer offline, after the training stage. During deployment, we extract features of the live observation and compute the cosine similarity with all the stored features in parallel. We justify this choice and compare it to several alternatives in Section \ref{sec:compare_retrieval}. This process is illustrated in Figure \ref{fig:retrieval}.

\label{sec:summary_framework}

As we will show in our experiments, retrieval from the buffer unlocks effective generalisation, while decomposition of object manipulation into alignment and replay brings a substantial improvement in learning efficiency. While the latter has been explored in \cite{johns2021coarse, valassakis2022demonstrate, di2022learning}, the combination of retrieval and decomposition is distinctly novel, and leads to a general and scalable framework that can continue to improve with further data. Furthermore, with this framework, teaching a task on an object takes \textit{no more than a minute of human time}, and this is an important property when considering practical, real-world imitation learning. The human operator simply provides just a single demonstration per object; the remainder of the data collection (the visual observations) is autonomous, and training is therefore self-supervised.

\section{Experiments}
\label{sec:experiments}
 
In this section, we empirically measure the ability of our \textit{retrieval, alignment, replay} framework to learn behaviours efficiently and transfer them effectively to new objects.
We use four everyday tasks: \textbf{grasping} an object, \textbf{pouring} from a cup into a container, \textbf{unscrewing} a bottle's cap, \textbf{inserting} a bottle's cap, and \textbf{inserting} toy bread in a toy toaster (we combine these two inserting tasks in one group). These tasks require non-trivial 6-DoF interactions with objects, and some require considerable precision (inserting a bottle's cap allows for a maximum of 4mm of error). We invite the reader to watch the videos on our website at \href{https://www.robot-learning.uk/retrieval-alignment-replay}{https://www.robot-learning.uk/retrieval-alignment-replay}. \\

\begin{figure}[t!]
    \centering
    \includegraphics[width=0.5\textwidth]{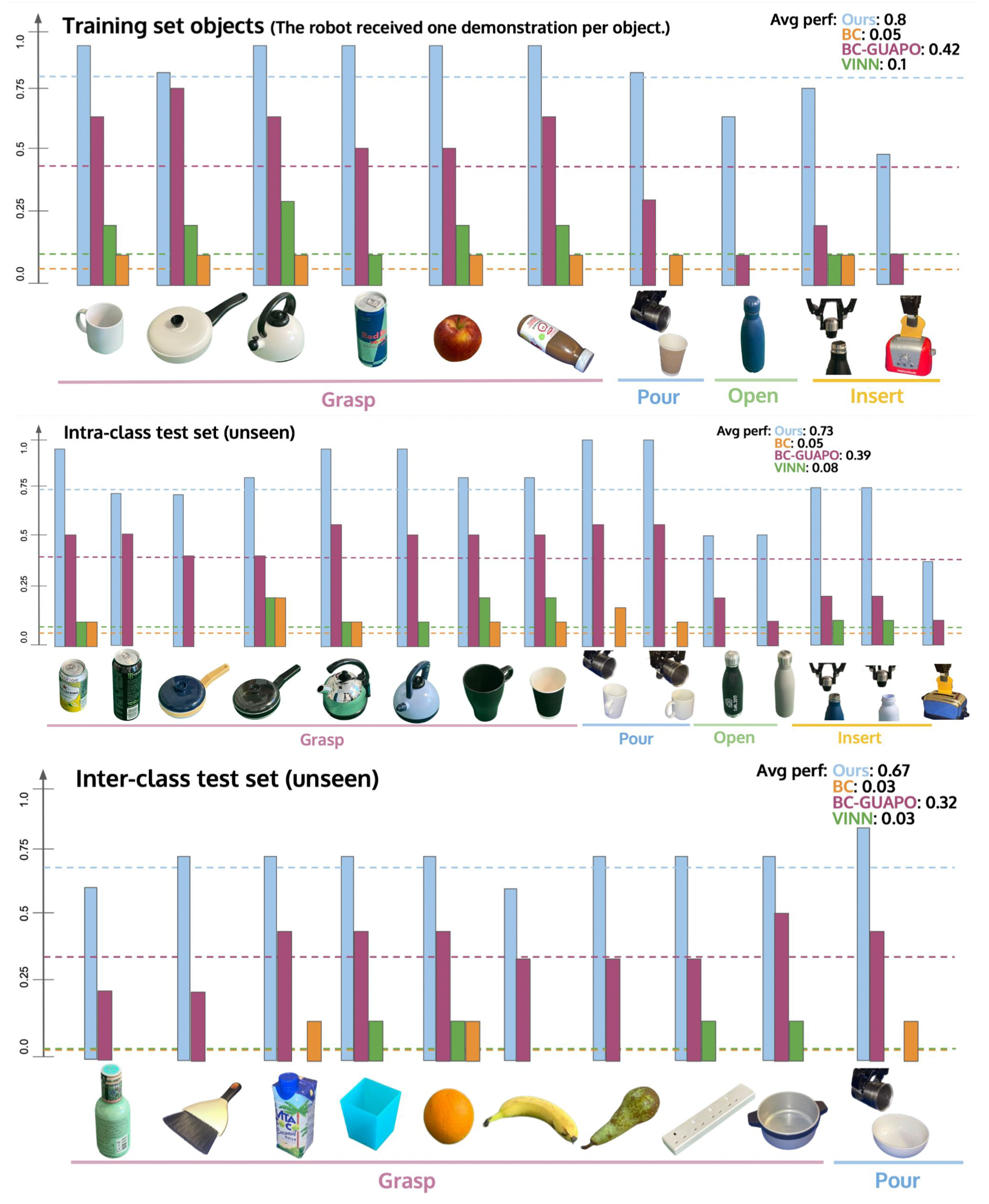}
    \caption{Performance of our method and baselines on a series of manipulation tasks involving tens of different objects, with the average performances as a dashed line.}
    \label{fig:all_testing}
    \vspace{-5pt}
\end{figure}

\textbf{Setup:}
We run our experiments on a Sawyer robot with a Robotiq gripper. We use a wrist-mounted RGBD Intel RealSense camera only, which does not need any calibration in our method. \revision{Our only assumption is that the end-effector to camera extrinsics are constant between training and testing, i.e. the camera remains fixed, a common assumption when mounting a wrist-camera to a robot \cite{borja2022affordance, di2022learning, johns2021coarse}.} The camera receives RGBD images which we rescale to $128\times128\times4$. We use a set of everyday objects (Figure \ref{fig:all_testing}) to illustrate the method's ability to tackle common tasks. The agent starts learning \textit{tabula rasa}: \textbf{no previous knowledge of tasks or objects, such as CAD models, is used}. Each method we study is trained with demonstrations using the objects depicted in Fig. \ref{fig:all_testing}, top bar plot. Each is then tested using the objects depicted in all three bar plots: training objects (top), intra-class unseen objects (middle), inter-class unseen objects (bottom). \revision{The wrist camera, rigidly mounted to the end-effector, starts the episode at 70cm above the table, and its field-of-view can observe the whole table in front of the robot, as demonstrated in a video on our website.}

\subsection{Baselines}
\label{sec:baselines}
Different methods in the recent literature have proposed augmenting robot manipulation with retrieval or decomposition of the trajectory, similar to the approach we adopt \cite{lee2020guided, wen2022you, borja2022affordance, pari2021surprising}. We investigate the contribution to manipulation performance brought by these building blocks, to motivate our design choices and better understand the effect of each.

\begin{figure}[t!]
    \centering
    \includegraphics[width=0.5\textwidth]{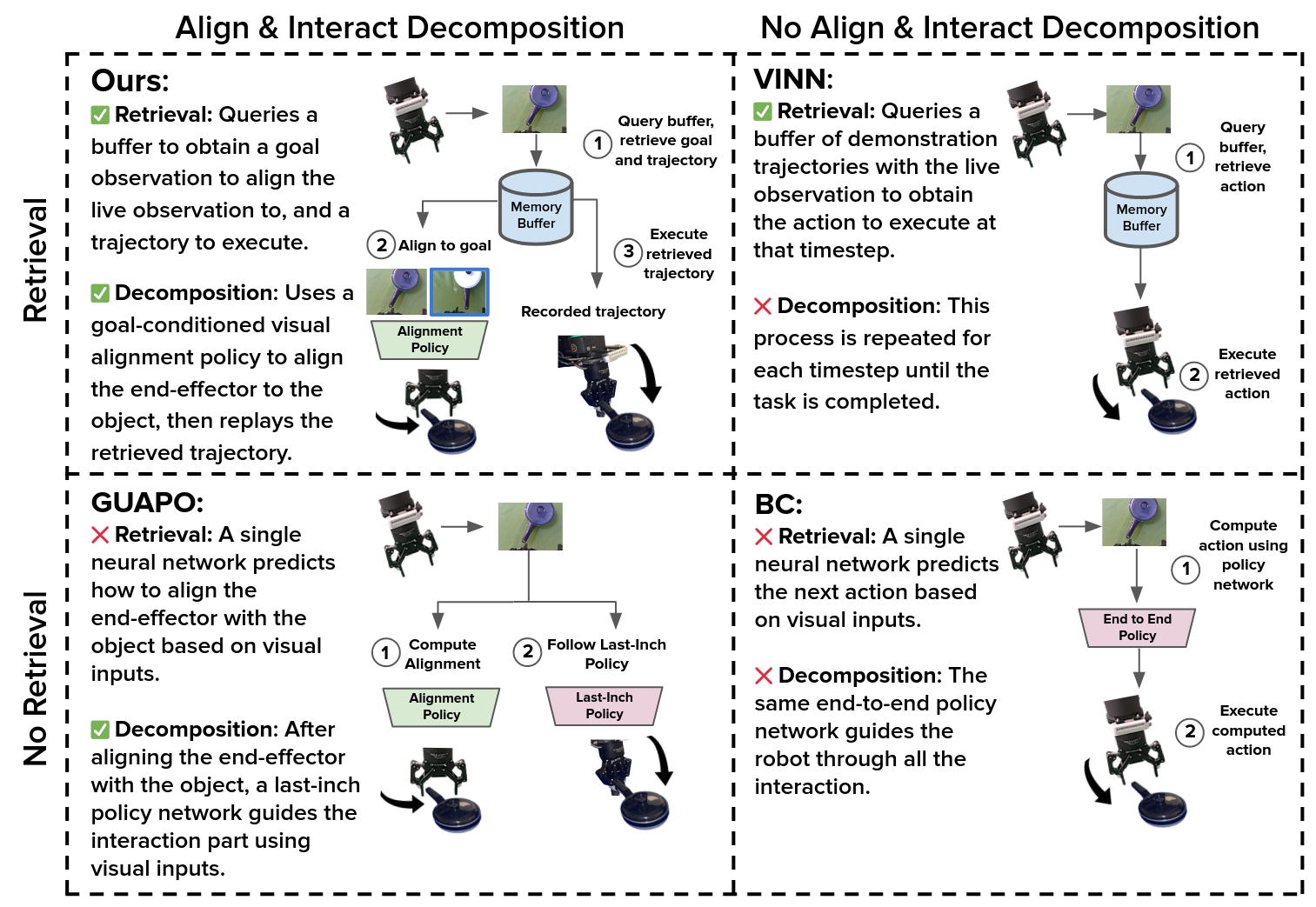}
    \caption{A taxonomy of the different robot manipulation paradigms we study and examples from recent literature, based on the concepts of retrieval and/or decomposition.}
    \label{fig:taxonomy}
\end{figure}

In particular, we identify 4 different families of approaches, grouped by whether or not they perform retrieval from a buffer of data gathered during demonstrations, and whether or not they decompose object manipulation into an alignment part followed by an interaction part, or treat it as a single learned trajectory. A visual taxonomy of these families is provided in Figure \ref{fig:taxonomy}. For each of those families, we implement and evaluate a method from the recent literature. To have a fair comparison, we chose methods that require similar assumptions to our method: as we discuss in Section \ref{sec:related}, some methods require either numerous, precisely calibrated cameras, and/or 3D models of the objects, while ours requires neither.

Our method is composed of both a \textit{retrieval} phase and a decomposition into \textit{alignment} and \textit{replay}. GUAPO \cite{lee2020guided} also decomposes object manipulation into a coarse \textit{alignment} phase, approaching the object by feeding visual observations into a neural network, and a last-inch learned policy is used to perform the interaction. This method lacks an explicit \textit{retrieval} phase, as the affordances are recognised implicitly by a neural network. To train this method, we use the same steps described in \ref{sec:where} to record data and train an alignment visual servoing network, which in this case is not goal conditioned, but only receives the live observation as an input.  Differently from our method, it uses a last-inch policy network that we train through Behaviour Cloning to interact with the object. As the original paper \cite{lee2020guided} proposes the use of Reinforcement Learning for the policy, while we use Behaviour Cloning for time efficiency, we denote this method as BC-GUAPO. 

Visual Imitation through Nearest Neighbours (VINN) \cite{pari2021surprising}, frames Learning from Demonstration entirely as a retrieval problem. Demonstrations are stored frame by frame in a memory buffer, together with the action taken by the expert at that timestep. At each timestep during deployment, the agent queries the buffer with the current observation to find the most similar $k$ observations. The action to execute is computed as a weighted average of the actions corresponding to these $k$ retrieved observations. While it is retrieval based, this method does not decompose object manipulation trajectories, and instead models manipulation as a single trajectory.

Finally, Behaviour Cloning (BC) \cite{open_x_embodiment_rt_x_2023} is implemented as a baseline that does not use retrieval nor decomposition, but trains a single policy network with the data collected during the demonstrations, and executes its actions at deployment guided by visual observations. In this case, we train an end-to-end policy network through Behaviour Cloning using the collected demonstrations.

We optimised neural network architectures and hyperparameters independently for each method, in an attempt to maximise the performance of each and obtain a fair comparison across these paradigms. To improve generalisation to visually similar objects, all the alignment or policy network we train for each method extracts visual features from the RGB channels through a pretrained, frozen ResNet-50 backbone \cite{he2016deep}, and then concatenates the depth channel, scaled to be in the same range as the extracted features.



\subsection{General Everyday Objects Interaction}
\label{sec:general_exp}

\textbf{Can our framework learn everyday-like manipulation behaviour efficiently, and transfer those trajectories to novel objects? What contribution do retrieval and decomposition bring, and how does this paradigm compare against recent baselines from the literature?}
To answer this, we train all baselines providing demonstrations of how to interact with the objects in the training set (Fig. \ref{fig:all_testing}, top). As our method requires a single demonstration per training object, in our first experiments we provide a single demonstration per training object to all the baselines as well.

We cannot measure inter-class performance on the \textit{bottle cap insertion}, \textit{unscrewing a bottle cap} and \textit{inserting toy bread in a toaster} tasks, as these tasks require a particular class of target object: we therefore only test for inter-class generalisation on the \textit{grasping} and \textit{pouring} tasks, that can instead be performed on a larger variety of unseen object classes (e.g. pouring can generalise from mugs to bowls). 
At test time, we present each object from both the training set and the test set to the robot. For each testing episode, the object is randomly positioned on the table in front of the robot. Through retrieval, the robot must understand how to interact with it based on the demonstrations it received (i.e. what task to perform: grasp it, open it, etc.). For example, if the robot observes a toaster and the bread slice in its end-effector, it should infer that the task is to \textit{insert the bread}. It must then correctly execute the task, hence measuring its ability to generalise to novel poses of the objects, and also to novel objects. We run 10 trials per object, sampling a new \revision{$(x,y,z,\theta_z)$} pose each time. 
\textbf{We train on 10 objects and test on 35 objects, for a total of 350 test-time episodes for each baseline. Of these 35 objects, 25 were novel objects, and 10 belonged to completely novel classes, demonstrating strong generalisation  and adaptation abilities.}

Results for our method's performance on each single object can be seen in Fig. \ref{fig:all_testing}. Here we show the success rate, and group those results into training set objects (for which the robot received demonstrations), intra-class objects (unseen objects from the same categories of the training objects, e.g. mugs), and finally inter-class objects (unseen objects from novel categories). \revision{All these results are obtained without the need for either intrinsic nor extrinsic camera calibration, and without any form of prior object knowledge}.

The results indicate not only that our method can obtain remarkable performance with just a single demonstration per object (an average of 80\% success on training objects after receiving a single demonstration for each), but that it can generalise to intra and inter-class objects: the average performance stays considerably close to the training set performance. (73\% success percentage on the intra-class set with respect training set, and 67\% on the inter-class set). 
\revision{The strong performance of our method on unseen, inter-class objects demonstrates that the proposed pipeline is robust even when there is not a clear object to retrieve from the buffer. When asked to grasp a banana (see Fig. \ref{fig:all_testing}), our method  understands that the most similar object is an horizontally-placed bottle by matching geometrical features, as both share a locally cylindrical-like shape, as we show in the videos on our website. While performance on unseen, inter-class objects is lower than on training or inter-class testing objects, we can observe a much less steep decrease compared to baselines that do not use retrieval and/or alignment (Fig. \ref{fig:all_testing}).}

\subsection{Investigating the Benefits of Retrieval and Alignment}
\label{sec:compare_decomp}

\begin{table*}[t!]
\footnotesize
\parbox{1.\linewidth}{
    \vspace{10pt}

\centering
    \caption{Success rate for trajectory replay and closed-loop learned policies on 6 tasks from our test set.}
    \label{table:how_table}
        \label{table:how_table}
        
    \begin{tabular}{cccccccc}
        \toprule
        {\textbf{Method / Task}} & Grasp Can & Grasp Mug & Unscrew Bottle & Pour & {Close Bottle} & {Insert Bread} & $ \textbf{Mean  ± Standard Deviation}$\\
        \midrule
        \textbf{Trajectory Replay (Ours)} &  $\textbf{0.8}$ & $0.7$ &  $\textbf{0.5}$ & $\textbf{0.9}$ & $\textbf{0.8}$ & $\textbf{0.7}$ & $ \textbf{0.73 ± 0.12}$ \\
        \textbf{BC-GUAPO 1} &  $0.3$ & $0.3$ &  $0.0$ & $0.5$ & $0.2$ & $0.1$ & $0.23  \text{ ± } 0.16$\\
        \textbf{BC-GUAPO 10} &  $0.7$ & $\textbf{0.8}$ &  $0.1$ & $0.8$ & $0.6$ & $\textbf{0.7}$ & $0.62  \text{ ± } 0.24$ \\
        
        \bottomrule

    \end{tabular}

}
\label{table:how_table}
\end{table*}

\begin{figure}[t!]
    \includegraphics[width=0.48\textwidth]{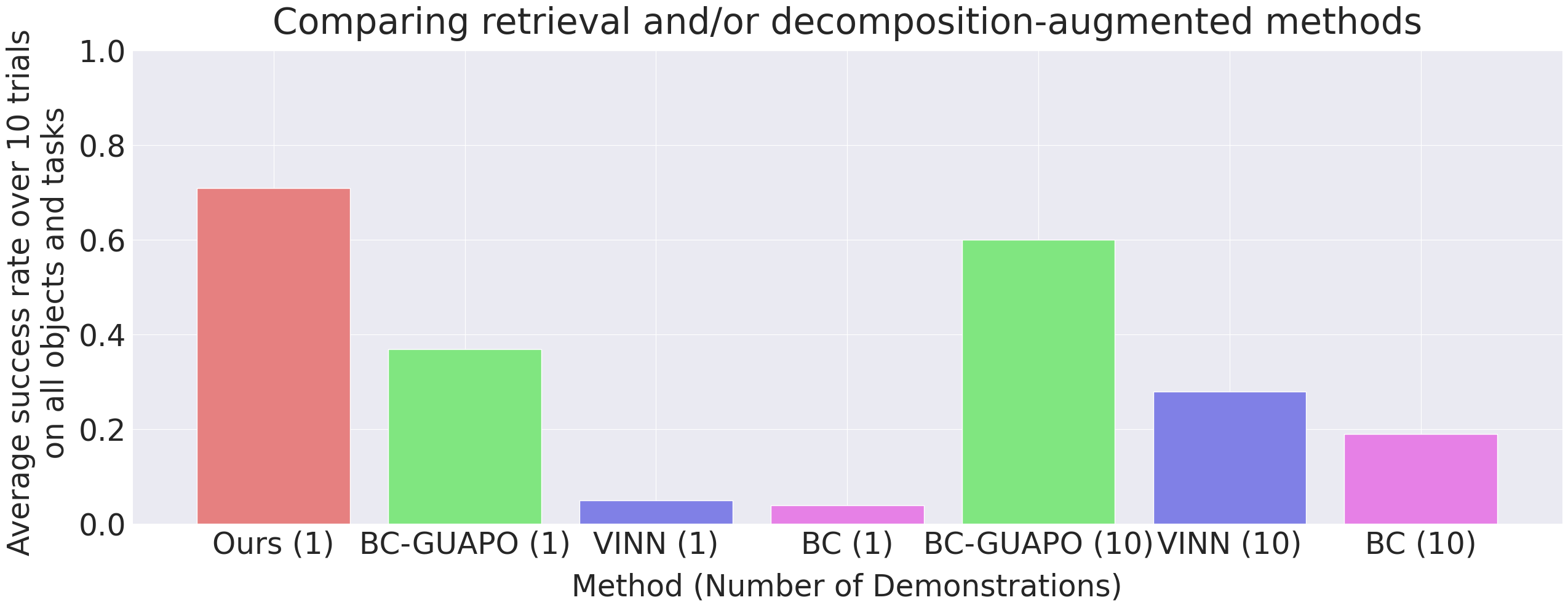}
    \caption{Average success rate of our method and the baselines, where each is provided with either (1) or (10) demos per training object.} 
    \label{fig:barplot}
  \vspace{-5pt}
\end{figure}

\textbf{Can we measure the performance benefits brought by retrieval and alignment both in isolation and then combined?}
We compare the overall performance of all methods, and how each scales with more data, in order to more closely study the individual contributions of the retrieval component, and the decomposition component (into alignment and replay). Figure \ref{fig:barplot} shows the results of these experiments, where each method is provided with either 1 demonstration or 10 demonstrations per training object, and where we measure the average success rate over all the objects and tasks of both the training set and the intra and inter-class test sets, running 10 episodes per object.

Comparing these results with the taxonomy proposed in Figure \ref{fig:taxonomy} allows us to clearly identify the contribution brought by the align/interact decomposition and/or retrieval. Behaviour Cloning uses neither of those, performs the worst of all the baselines whether 1 demonstration or 10 demonstrations per object are provided. VINN is based exclusively on retrieval, and when receiving a single demonstration per object performs poorly as well. This suggests that retrieval alone is not sufficient to enable efficiency and effective transfer. On the other hand, BC-GUAPO, which decomposes trajectories into two phases similar to our method, performs much better than VINN with just a single demonstration per object. This clearly indicates the benefits of decomposing manipulation in this way, because the coarse alignment phase can be trained in a self-supervised manner from just a single demonstration.

When comparing BC-GUAPO to our method, however, we see a substantial gap in performance: as our method is based on both retrieval and decomposition, this suggests that \textit{retrieval brings a noticeable performance boost when paired with decomposition}. In particular, from our experiments it emerges that retrieval benefits from a large memory buffer of observations: when receiving 10 demonstrations per object, VINN clearly surpasses BC as well due to VINN benefitting from the use of retrieval. Our framework combines the best of both worlds: the decomposition into \textit{alignment} and \textit{replay} enables the robot to autonomously gather a large buffer of demonstrations without the need for human supervision. Given this buffer, the use of \textit{retrieval} then enables the robot to generalise efficiently to novel objects.

\subsection{A Comparison of Trajectory Replay and Learned Policies}
\label{sec:compare_replay}
\textbf{Is replaying a recorded trajectory as effective as learning a policy and executing it in closed-loop guided by visual inputs?} At a first glance, replaying a trajectory may sound suboptimal. But as demonstrated by recent works \cite{johns2021coarse, valassakis2022demonstrate, di2022learning}, if the alignment is precise, replaying such trajectories can obtain equal if not superior performance, in addition to being substantially more efficient than training a policy function in the form of a neural network.

To highlight this effect, we conduct an experiment that compares our replay method to a baseline that uses a policy network, where both begin after aligning the end-effector with the object. We ignore the retrieval phase here, and train and test each task in isolation. First, we collected results for our method without the retrieval phase, using only the alignment and replay phases after manually selecting the task to execute (as a goal image and trajectory), which is thus very similar to C2F-IL \cite{johns2021coarse}. Then, we collected results for our BC-GUAPO baseline. Both methods share the same alignment policy, trained as in Section \ref{sec:method}, and the different interaction phases are hence the only source of difference in performance: once the alignment is complete, our method executes the recorded trajectory, while BC-GUAPO uses a learned policy network, feeding it visual inputs from the wrist-camera.

We train the two methods on a subset of the tasks: two grasping tasks, unscrewing a bottle, closing a bottle, pouring from a cup into a container, and inserting toy bread into a toy toaster (Table I). We provide a single demonstration per task to our method and to BC-GUAPO 1, while we provide 10 demonstrations to BC-GUAPO 10. We then test the performance on an unseen object of the same class per each task and class. We show results in Table I. 
This clearly proves the effectiveness of trajectory replay compared to explicitly learning an end-to-end policy, for a range of everyday tasks, beyond just grasping. As demonstrated in \cite{open_x_embodiment_rt_x_2023}, an end-to-end policy network may need hundreds of demonstrations per task to correctly learn skilful manipulation of everyday objects, whereas our experiments show that \textit{trajectory replay is a simple but effective approach when data efficiency is important}.

\subsection{A Comparison of Retrieval Techniques}
\label{sec:compare_retrieval}
\textbf{What technique performs best in retrieving the most similar observation from the memory buffer?} We compare the of use pre-trained neural networks trained on large datasets of observations, including both those which do and those which do not have a robotics focus. The process can be formulated as follows: given an observation $o_i$, we compute a representation by feeding it into a neural network, and extracting either an output embedding vector, or an intermediate representation. We use ResNet-50 \cite{he2016deep}, CLIP \cite{radford2021learning}, R3M \cite{nair2022r3m}, and ViT DINO \cite{caron2021emerging}.\\

 \begin{figure}[t!]
    \includegraphics[width=0.48\textwidth]{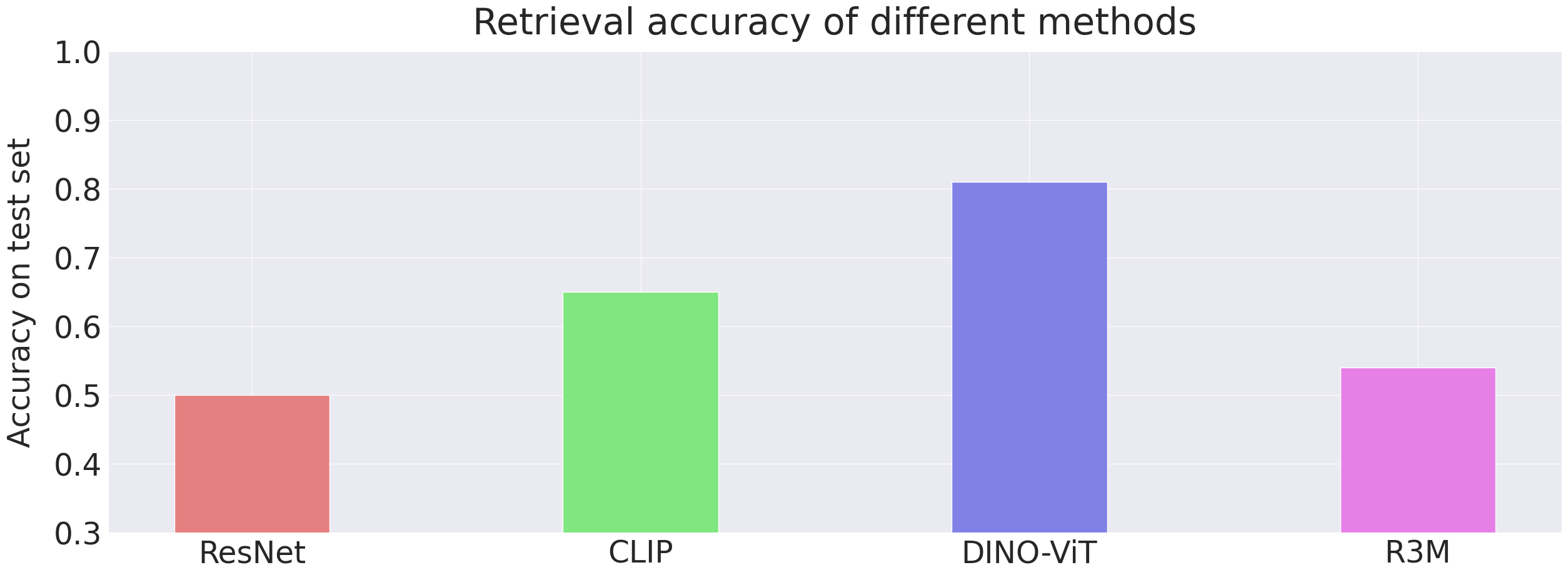}
    \caption{Average retrieval accuracy of different feature extraction methods. We test each method on the intra-class test set (Fig. \ref{fig:all_testing}, middle), measuring the ability to retrieve objects of the same class.}
    \label{fig:retr_accuracy}
    \vspace{-10pt}

\end{figure}

We measure retrieval accuracy by providing to each method an observation of an unseen object from the intra-class test set (e.g. unseen mugs, pans, teapots). We then measure how many times the method retrieves from the buffer an observation of the object belonging to the same class, indicating that the representation computed from that observation maximises the similarity of the representation computed on the test object observation. We show results in Figure \ref{fig:retr_accuracy}. The features extracted via a Vision Transformer trained with the DINO framework achieve the best accuracy, confirming recent findings from the literature \cite{caron2021emerging, amir2021deep} that such features encode semantic, geometric, global and local information of the observation effectively. 



\section{Discussion}
\label{sec:final_discussion}


\subsection{Further Questions and Limitations} \textbf{What if there are more objects at test time?} Currently, our method relies on visual inputs to predict what to do in an environment: multiple objects may therefore lead to ambiguity. However, if we use an off-the-shelf object segmentation method, we can in principle interact with each object in succession. Future work will address these limitations. \textbf{Why limit the alignment phase to 4-DoF?} \revision{This is not a fundamental limitation of our method; if required, the alignment phase could be extended to 6-DoF by simply collecting images over 6-DoF end-effector poses. Also, in our tabletop implementation, although the alignment phase is 4-DoF, the replay phase is in fact 6-DoF, enabling complex interactions}. \textbf{What about tasks that require feedback loops, pushing objects to a target?} Our method cannot be applied to these methods in its current form, but in this work we show that a large set of everyday-tasks can be tackled with our framework, and that it is surprisingly competitive and often better in terms of efficiency and performance compared to learned closed-loop policies.  \textbf{What happens if you want to interact with the same object in different ways?} If an object can be interacted with in multiple ways, the robot needs an external input, provided by e.g. a planner or the user themselves, to select an interaction. Given this, our proposed framework could then still be used for executing an object manipulation task.

\subsection{Conclusions}

In this work, we propose a novel framework to enable efficient and effective learning from demonstration in robotic manipulation. In particular, we introduce a combined use of \textit{retrieval}, and trajectory decomposition into \textit{alignment} and \textit{replay}. We analysed the best way of implementing retrieval and decomposition by comparing baselines from the literature, and showed that our combination of retrieval, alignment, and replay, performs better than alternative paradigms. Furthermore, we demonstrated how retrieving and replaying demonstration trajectories is a simple yet powerful alternative to training end-to-end policies, with significant gains in data efficiency. 

\bibliographystyle{plainnat}
\bibliography{main}

\section{Supplementary Material}

\subsection{Algorithms}
Here we describe our method in algorithmic form, starting from demo and data collection in Alg. \ref{alg:demo}, then describing the deployment phase in Alg. \ref{alg:deployment}.

\begin{algorithm}
\caption{Pseudo code for data collection and demo recording.}
\label{alg:robot_navigation}
\begin{algorithmic}
\FOR {n in objects} 
\STATE Operator brings robot to $b^W$ 
\STATE $o_{b} \leftarrow$ Record observation
\STATE demo\_buffer $\leftarrow$ Store $o_{b}$
\FOR{$t$ in $\text{range}(T)$}
\STATE Go to random pose $p^W_t$
\STATE $o_{n,t} \leftarrow$ Record observation
\STATE $d^W_t \leftarrow b^W_t - p^W_t$
\STATE $d^E_t \leftarrow {^E}T_W d^W_t$
\STATE demo\_buffer $\leftarrow$ Store $o_{n,t} , \text{ }d^E_t$
\ENDFOR
\STATE Go to $b^W$
\STATE trajectory $\leftarrow$ Operator records demo from $b^W$
\STATE demo\_buffer $\leftarrow$ Store demo
\STATE full\_buffer $\leftarrow$ demo\_buffer 
\ENDFOR
\end{algorithmic}
\label{alg:demo}
\end{algorithm}

\begin{algorithm}
\caption{Pseudo code for object interaction at test time.}
\begin{algorithmic}
\STATE $\gamma = 5 \times 10^{-3}$m 
\STATE $d^E_t = \infty$ 
\WHILE {$d^E_t > \gamma$} 
\STATE $o_{live} \leftarrow$ Record observation
\STATE $o_{j,n} \leftarrow \text{argmax}_{o_{i,d}} g(o_{live},o_{i,d})$ Extract most similar obs from buffer
\STATE $o_{b,n} \leftarrow$ Retrieve corresponding bottleneck obs from $o_{j,n}$  
\STATE $d^E_t \leftarrow f_\theta(o_{live}, o_{b,n})$ Compute displacement to align $o_{live}$ with $o_{b,n}$
\STATE Execute $d^E_t$
\ENDWHILE
\STATE trajectory$_n$ $\leftarrow$ Retrieve corresponding demo from $o_{j,n}$
\STATE Execute trajectory$_n$

\end{algorithmic}
\label{alg:deployment}
\end{algorithm}

\subsection{Test Time Examples}
We provide here a series of frames taken from one of the videos on our website, showing the robot tackling a grasping, pouring, inserting and unscrewing. More videos are available at \hyperlink{https://www.robot-learning.uk/retrieval-alignment-replay}{https://www.robot-learning.uk/retrieval-alignment-replay}.

 \begin{figure}[b!]
    \centering
    \includegraphics[width=0.45\textwidth]{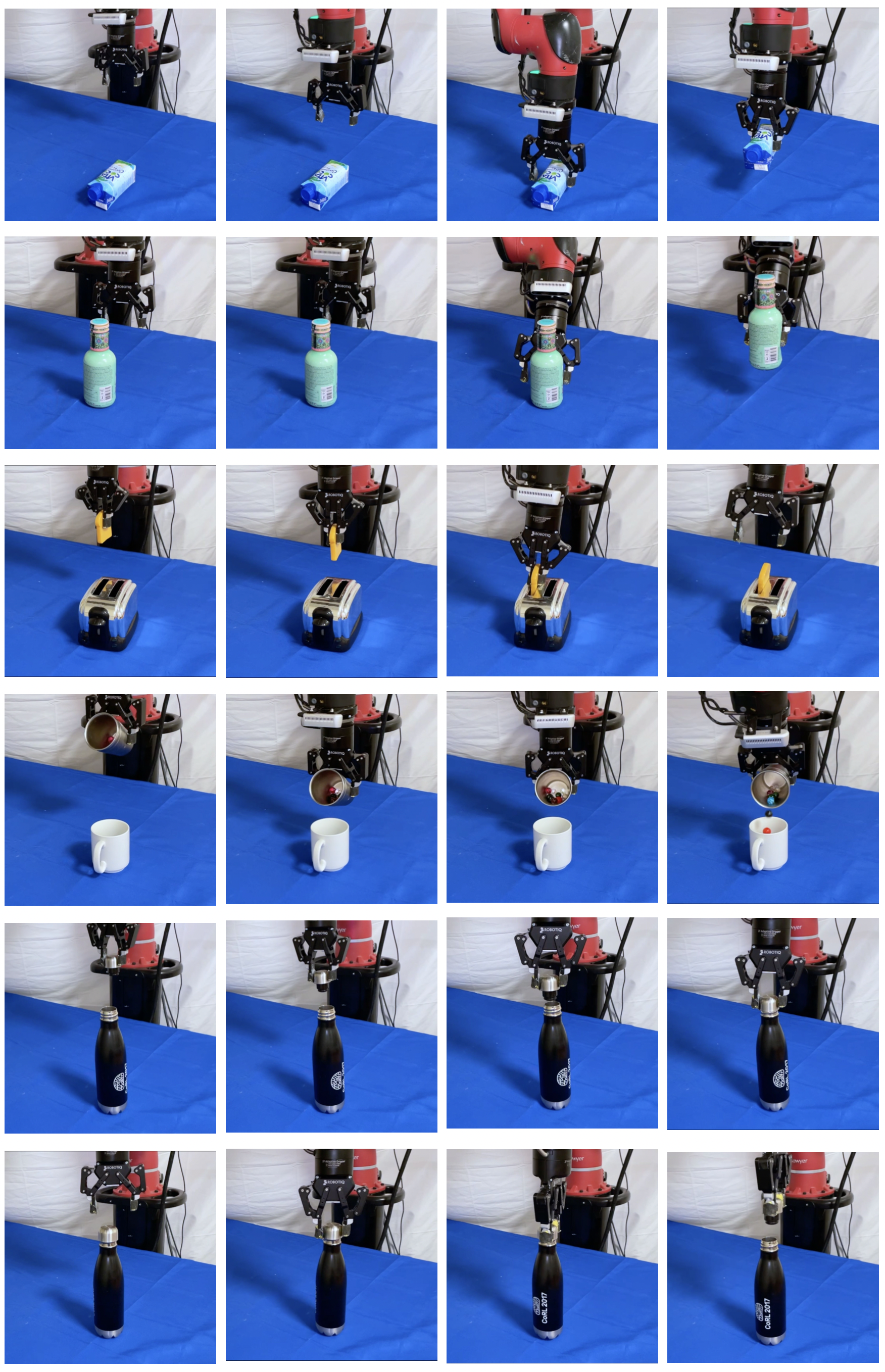}
    \caption{A series of frames from one of our videos.}
    \label{fig:full_results}
\end{figure}


\newpage

\end{document}